\documentclass{article}

\usepackage[preprint]{neurips_2021}

\usepackage[utf8]{inputenc} 
\usepackage[T1]{fontenc}    
\usepackage{hyperref}       
\usepackage{url}            
\usepackage{booktabs}       
\usepackage{amsfonts}       
\usepackage{nicefrac}       
\usepackage{microtype}      
\usepackage{xcolor}         
\usepackage{graphicx}
\usepackage{multirow}

\setcitestyle{authoryear, open={(},close={)}}

\title{Multilingual Byte2Speech Models for Scalable Low-resource Speech Synthesis}

\author{%
  Mutian He$^1~~~~~~~~$ 
  Jingzhou Yang$^2~~~~~~~~$
  Lei He$^2~~~~~~~~$
  Frank K. Soong$^2$  \\
  \\
  $^1$ 
  The Hong Kong University of Science and Technology\\
  $^2$ 
  Microsoft China\\
  \texttt{mhear@cse.ust.hk}
  \\
  \texttt{\{jingy,helei,frankkps\}@microsoft.com}
}

\begin{document}

\maketitle

\begin{abstract}
To scale neural speech synthesis to various real-world languages, we present a multilingual end-to-end framework that maps byte inputs to spectrograms, thus allowing arbitrary input scripts. Besides strong results on 40+ languages, the framework demonstrates capabilities to adapt to new languages under extreme low-resource and even few-shot scenarios of merely 40s transcribed recording, without the need of per-language resources like lexicon, extra corpus, auxiliary models, or linguistic expertise, thus ensuring scalability. While it retains satisfactory intelligibility and naturalness matching rich-resource models. Exhaustive comparative and ablation studies are performed to reveal the potential of the framework for low-resource languages. Furthermore, we propose a novel method to extract language-specific sub-networks in a multilingual model for a better understanding of its mechanism.
\end{abstract}

\section{Introduction}

Recent years witnessed the magnificent triumph of end-to-end deep learning. 
Particularly for speech synthesis (or text-to-speech, TTS), pipelines and handcrafted features are substituted by end-to-end neural models \citep{DBLP:conf/icassp/ShenPWSJYCZWRSA18Tacotron2}. 
But such a success relies on a rich resource of high-quality data, 
and data requirements have thus become a bottleneck of research and application on neural TTS. Various researches have been proposed for the issue. However, most of them require extra resources. For instance, phoneme-input models reduce the modeling complexity and reach better performance \citep{DBLP:journals/csl/YasudaWY21}, 
but lexicons and/or grapheme-to-phoneme (G2P) rules of the target language must be given. 
While character-input methods require knowledge of the script to define model inputs on non-alphabetic scripts such as Indic and Chinese ones.
Some methods leverage phonology knowledge to reduce data requirements \citep{DBLP:journals/corr/abs-2005-10441,DBLP:conf/interspeech/ChenTYL19, DBLP:conf/sltu/DemirsahinJG18}, thus linguistic expertise on the target is essential. Besides, a speech chain of synthesis and recognition plays a key role in the low-resource TTS by \citet{DBLP:conf/icml/RenTQZZL19} and \citet{DBLP:conf/kdd/XuTRQLZL20LR}, which relies on an additional large paired corpus to train a recognizer. To conclude, all such methods depend on extra resources: data, knowledge, or developer's efforts for \textit{each} target language.

It is often costly to prepare the resources and build a complex system on a language, not to say to repeat this on thousands of languages in the world.
Therefore, we propose a novel scalability-centered task: We must refrain from using extra per-language resources. Developers should not put efforts on any language, but rely on paired TTS data which often could be collected in a standardized process. 

For such a challenge, we highlight transfer learning from rich resource languages. 
Performances of low-resource languages are improved in a multilingual model on a cohort of languages \citep{DBLP:conf/interspeech/LiZ16,DBLP:conf/sltu/DemirsahinJG18,DBLP:conf/interspeech/KorteKK20,DBLP:conf/interspeech/YangH20}. Despite the immense diversity of languages, identical or similar writing systems, G2P rules, phoneme inventory, or at least the pattern to learn sequential mapping in TTS can be leveraged for transfer learning.
Particularly, we extend \citet{DBLP:conf/icassp/LiZSWC19}'s method.
By leveraging the existent text encodings of UTF-8, arbitrary scripts covered by Unicode are supported without our need to study individual writing systems. We build a strong multilingual multi-speaker transformer on a 900-hour corpus of 43 languages by 109 speakers written in various scripts. 
Using a \textit{tier-wise progressive} and language-balanced strategy, such a model learns to produce correct and natural speech similar to phoneme-based methods. 

We then evaluate its capability to adapt to selected low-resource target languages upon a linguistic basis. Through a \textit{co-training} strategy, it can learn a brand new language such as Romanian and Greek in a few-shot regime of 10 samples or less than 1 minute of audio, and to reach topline performance with much fewer data. In this way, more marginalized, endangered or underrepresented languages may benefit from neural TTS. Also, we investigate the contribution of various factors in our framework by exhaustive empirical studies. 
Furthermore, to better understand the mechanism of the model, we propose a novel approach to interpret a multilingual model by language-specific pruning and show that a multilingual model can be viewed as a fusion of monolingual models sharing part of parameters between each other, which \textit{emerges} from training.
To facilitate reproduction, we present a pipeline based on open resources.\footnote{The pipeline and audio demos are available at \url{github.com/mutiann/few-shot-transformer-tts}.} To conclude, the contributions of the paper are three-fold:

\textbullet\ We build a multilingual Byte2Speech framework for a scalability-centered TTS task. With well-designed training strategies, it matches performance of phoneme models on a variety of languages.

\textbullet\ We investigate its low-resource capabilities to adapt to new languages in few shots and to reach results similar to baseline models using data reduced by an order of magnitude.

\textbullet\ We deepen our understandings of multilingual model mechanisms by a novel interpretation.

\section{Methods}
\subsection{Framework}
We adopt the multilingual and multispeaker transformer TTS framework in \citet{DBLP:conf/interspeech/YangH20} and extend it to byte inputs. In detail, we train a 12-layer transformer to predict mel-spectrograms, with language and speaker embeddings concatenated by encoder outputs, while we find no improvements if we feed language ID to inputs.
Inputs texts are encoded in UTF-8, each byte as a token, and fed into the model along with [BOS] and [EOS], so the vocabulary sizes 256 except for special tokens like [BOS]. 
Hence, in our inputs, a character can be represented in one or more tokens.
We directly use encoded texts for training, with only basic pre-processings like rule-based transformations of digits and Chinese word segmentation. Besides, Unicode characters with separable parts can often be represented in multiple ways. 
For example, 
each precomposed Korean character stands for a syllable, 
but it is equivalent to record it by a series of \textit{Jamo} (letters). Similar cases include ligatures and letters with diacritic marks.
Although using non-precomposed tokens reveals more about a character, following our principle for using minimal language-specific knowledge, we do not apply any normalization on characters and allow precomposed characters to present in data as-is.

\begin{figure}[t]
\begin{center}
\centerline{\includegraphics[width=0.8\columnwidth]{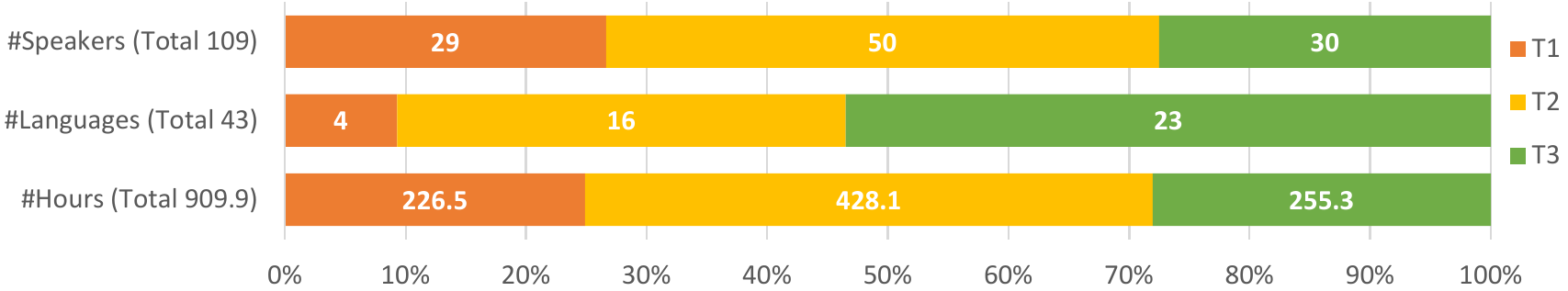}}
\caption{Statistics of the training corpus, as for amounts of data, languages, and speakers per tier.}
\label{tiers}
\end{center}
\vskip -0.2in
\end{figure}

\subsection{Training corpus}
We use 43 languages as our \textit{source} languages, with 32 distinct languages in ISO-639-1, plus their regional variants.\footnote{For simplicity we use the term “language” for all of them. See Appendix Section~C for dataset details.} The corpus thus covers a majority of human scripts and phonemes. 
Particularly, besides Latin alphabets, there are also Cyrillic alphabets, Chinese characters, 
syllabic scripts (like Korean), abugida (like Hindi), and abjad (like Arabics) .
With such diversity, relationships between text and speech are highly disparate. Languages like Spanish use a phonemic orthography, while English G2P is far from regular. Alphabets record all phonemes, while abjads typically omit vowels. Chinese logograms do not record pronunciations, and in Japanese, a Chinese character can have 10 different readings. 
Making it worse, as in the real world, language resources are highly imbalanced. We split the languages into three \textit{tiers} by number of samples. As shown in Figure~\ref{tiers}, T1 languages like English (US)
possess a disproportional number of samples and speakers, while only $<$10k samples are available for each T3 language. Therefore, despite the large corpus, obtaining a multilingual source model remains challenging.

\subsection{Adaptation targets}

To best reveal the adaptation capabilities, we inspect the linguistic similarity between each adaptation target and each tier of source languages, including aspects of writing systems, pronunciation or G2P rules, lexicon, and phonetic traits. Based on this, we pick five particular targets:

\textbullet \- Indian English (en-in), which, as for our data, is mostly mutual-intelligible with other English variants as for writings, G2P, lexicon, and phoneme inventory. However, there are many phonetic shifts, such as pronouncing “th” in “then” as voiced dental plosive similar to /d/.

\textbullet \- Romanian (ro-ro), which uses the Latin alphabet with five extra letters, and has a close-to-phonemic orthography that G2P is mostly one-to-one. The script, G2P, lexicon, and phoneme inventory are close to other Romance languages like Italian and French in T1, thus allowing an effective transfer.

\textbullet \- Greek (el-gr), which uses the Greek alphabet unseen in sources, thus the model needs to learn a new alphabet with each letter in two bytes. Greek orthography is of intermediate-depth with one-to-N relations from phoneme to letters, but G2P is mostly regular using rules similar to other alphabets. Also, there are lexical and phonetic links with other European languages, giving chance to transfer.

\textbullet \- Thai (th-th), which is written in 3-byte abugida unseen in sources. Abugidas, unlike alphabets, first record a syllable and then modify its vowel by diacritics. Only in T3 there are Indic languages using abugidas, but with different encodings. 
Furthermore, Thai is tonal, uses less phonemic G2P with rich irregularities, and has no particularly similar source language. All of these hinder adaptation.

\textbullet \- Mandarin Chinese (zh-cn) is a tonal language written in 3-byte simplified Chinese logograms, hence a byte-based model must memorize the reading of each character. Though some characters overlap with Japanese in T1 and Cantonese in T2 (in traditional Chinese), the spoken forms are systematically different. Thus it is particularly challenging and the help of source languages is dubious.


Besides, we analyze the phoneme inventory involved. The training corpus covers all phonemes used in target languages since T2, showing the capability of the source model to articulate the targets. Thus the key for adaptation is to learn to handle input scripts and pronunciation rules. While for lower tiers like T1, there is one phoneme from ro-ro absent, one from el-gr, two from th-th, and four from zh-cn. Those phonemes rare in source languages add to difficulties. To conclude, from en-in to th-th/zh-cn, the task becomes more challenging and similar source languages are present only in higher tiers.

\subsection{Training strategy}
Transformer TTS suffers from unstable training, often requiring tricks such as alignment constraints \citep{DBLP:conf/interspeech/ChenTRXSZQ20}. Byte inputs and multilingualism add to the complexity, making our \textbf{tier-wise progressive training} essential: We initialize all models by training on selected short en-us samples for 30k steps. Starting from it, T1 data are added, and then T2 data at total 350k steps, and T3 at 650k. 
Languages are exponentially balanced: For language $i$ with $N_i$ samples, we compute
\[ c_i=N_i / \sum_j {N_j} \]
\[ p_i=c_i^\alpha / \sum_j {c_j^\alpha} \]

During training, we first sample a language $i$ with probability $p_i$ using $\alpha=0.2$, and then a training example from it \citep{DBLP:conf/interspeech/YangH20}.
To improve efficiency, we use 4 GPUs with dynamic batching \citep{DBLP:conf/aaai/Li0LZL19}. Alternative training settings lead to suboptimal performance in experiments.

As for low resource adaptation, we discover that merely training by the target results in overfitting, and rich source languages can serve as regularization. Therefore, we adopt a multitask or \textbf{co-training} strategy. Instead of fine-tuning a well-pretrained model, we add the target language with $p_i=0.25$ to train with a mixture of sources and the target. To identify the impact of source languages, we attempt to 
perform adaptation after each tier transition plus using en-us only, that is to co-train with en-us from 30k steps, T1 from 350k, added by T2 from 500k, and T3 from 700k. Exponential learning rate decay is applied in a period of 850k steps, with the step counter reset at adaptation and tier transitions.

\subsection{Evaluation metrics}
For each language, we generate mel-spectrograms on a held-out set of 100 samples (utterances), in comparison with mels from recordings.
Waveforms produced by Griffin-Lim are sent to Azure Speech-To-Text to get character error rate (CER) as a large-scale intelligibility metric.
Also, we evaluate the quality by mean square error (MSE) with ground truth mels, both after collapsing unvoiced parts and dynamic time warping using FastDTW \citep{salvador2007toward}. 
Besides, we collect an extra set of news scripts of much more ($\sim$1000) samples per language with longer and more complicated sentences. We compute CER upon these scripts to build a \textit{CER-Ex} metric.
We further perform subjective tests on selected models. For intelligibility, we invite five judges per sample to annotate word errors on audios from the 100-sample held-out set. For mean opinion score (MOS), we invite 20 judges per sample on a 20-sample evaluation set selected from the held-out set, on which none of the models have mispronunciations, in order to best compare the naturalness. We use a pretrained WaveNet to generate waveform for subjective tests. Results are also compared with phoneme-based multilingual models by \citet{DBLP:conf/interspeech/YangH20} trained on the same dataset as a topline.

\section{Experiments}

\subsection{Source languages}
We first demonstrate intelligibility by CER as in Table~\ref{source}. The Byte (T3) model achieves CER comparable to the phoneme-based model on languages using phonetic scripts, including en-us from T1, German (de-de) from T2, and Telugu (te-in) from T3. These results show that a sophisticated byte-based model suffices to learn various phonetic scripts at the same time, and may even produce fewer errors than a phoneme-based model, possibly thanks to our training strategy and representation sharing of input tokens. This also applies to Telugu with fewer data and a unique abugida. As for the logographic Cantonese (zh-hk), CERs are both high due to rich homophones, but the gap between models is limited, showing that the byte model may memorize and reproduce the pronunciations of thousands of Chinese characters sparsely scattered in data. Therefore, our Byte2Speech model may reach competitive intelligibility on rich-resource languages without prior G2P knowledge.

\begin{table}[t]
\caption{CER (\%) on sources with data size labeled, using best results of each model}
\label{source}
\vskip 0.15in
\begin{center}
\begin{small}
\begin{sc}
\begin{tabular}{lcccc}
\toprule
Language & en-us (150h) & de-de (30h) & zh-hk (30h) & te-in (5h)\\
\midrule
Phoneme & 3.23 & 2.13 & 13.16 & 11.35\\
Byte (T3) 
& 2.43 & 1.18 & 15.67 & 9.58\\
\bottomrule
\end{tabular}
\end{sc}
\end{small}
\end{center}
\vskip -0.1in
\end{table}

\begin{figure*}[t]
\centerline{\includegraphics[width=\textwidth]{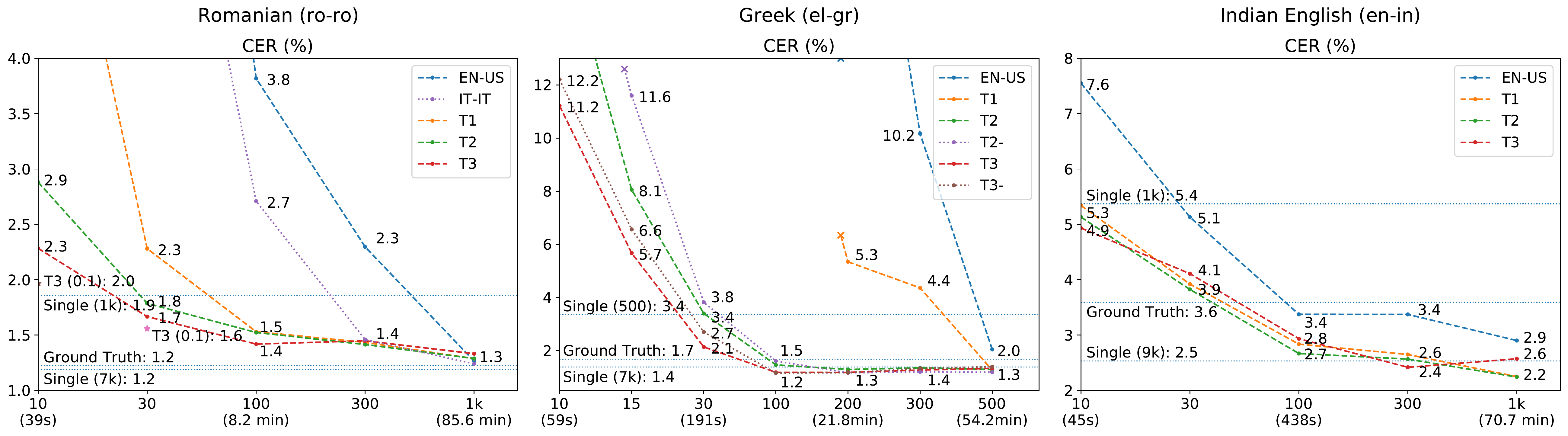}}
\caption{CER for Indian English (en-in), Romanian (ro-ro), and Greek (el-gr), with x-axis the number of samples or lengths of audio. For clarity, only one number is labeled for overlapping datapoints, and those too large are omitted. Crosses (×) represent failed models. Horizontal lines show ground truth and single-language model results with dataset sizes labeled. Isolated datapoints and curves labeled as IT-IT, T2-, and T3- will be discussed in Section~\ref{ablation}.}
\label{plot1}
\vskip -0.1in
\end{figure*}

\begin{figure*}[t]
\centerline{\includegraphics[width=\textwidth]{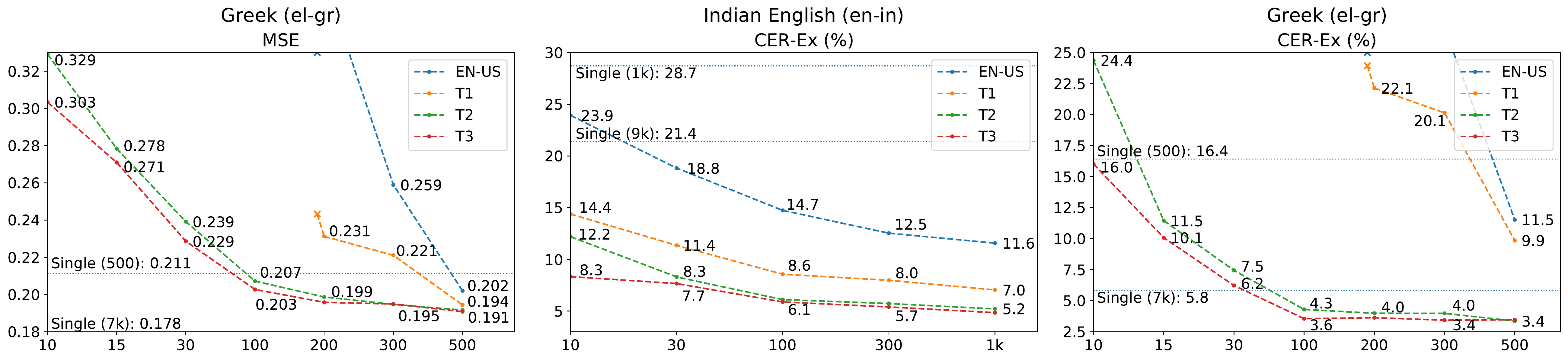}}
\caption{MSE for Greek (el-gr), and CER-Ex for Indian English (en-in) and Greek (el-gr).}
\label{plot1_ex}
\vskip -0.1in
\end{figure*}

\subsection{Adaptation}

We perform adaptation on targets with randomly sampled sub-datasets of different sizes, co-trained from different sources, and compare them with single-language models and recording mel CERs.

Romanian results in Figure~\ref{plot1} are the most representative: as shown by the T3 curve, with merely 10 samples or 39s recordings, the model acquires a brand new language of Romanian with high intelligibility of 2.3\% CER, entering the regime of few-shot learning. Besides, with 1k samples, adapted models reach CER close to ground truth and the full-data ($>$7k samples) single-language model, 30\% better than the 1k-sample single-language model.
Thanks to the co-training strategy, this even applies to adaptation from en-us on 1k samples.
But with fewer data, multilingualism becomes essential: With higher tiers or more source languages, all the metrics get constantly improved, especially with $<$ 300 samples.

More remarkably, even for a brand new alphabet of Greek, the model learns most pronunciations with 10 samples; and it easily reaches CER of 2.1\%, close to ground truth and the full-data model, using 30 samples. Besides, on Greek, the model may produce mels close to ground truth using much fewer data, as indicated by MSE in Figure~\ref{plot1_ex}. A diverse source corpus proves critical as en-us and T1 models fail to produce any intelligible speech when the data size is reduced to 200 or 100, respectively.

Indian English is simpler and closer to an English dialect, so all models reach a low CER even with few shots, and differences between T1/T2/T3 are limited. However, the CER gap between en-us and T1 still shows the impact of multilingualism. More, as shown in Figure~\ref{plot1_ex}, it is found in all languages but most remarkable in en-in that with a higher tier, adapted models have better CER-Ex, often beyond full-data (9k) models. On complicated texts in the CER-Ex test set, single-language models often produce mispronunciations and misalignments. Thanks to adaptation from a rich source corpus with more diverse scripts, the adapted model may generalize better to those difficult inputs. 


\begin{figure*}[t]
\centerline{\includegraphics[width=\textwidth]{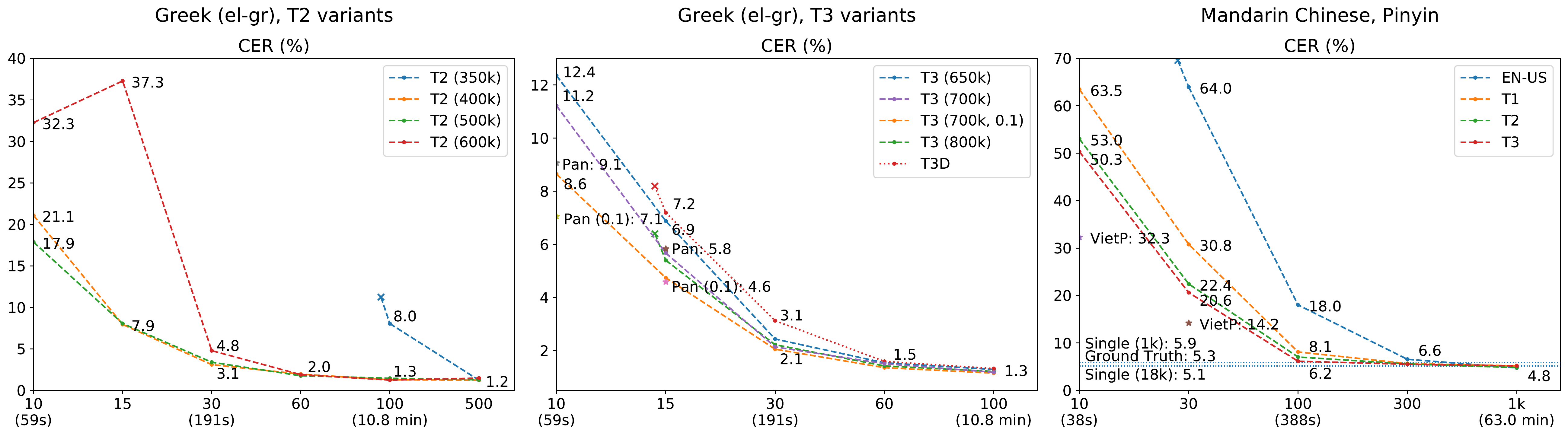}}
\caption{CER for alternative settings in Greek (el-gr) and Mandarin Chinese (zh-cn).}
\label{plot3}
\vskip -0.1in
\end{figure*}

The task is harder for Thai and Mandarin Chinese with less phonemic scripts. 100 Thai samples are required to reach 17\% CER, and 1k samples (71.4 minutes) to match the ground truth CER of 2.3\%. Mandarin Chinese is the most difficult, but a 2k-sample adaptation still reaches 9.2\% CER.
Although it seems formidable to acquire Thai or Mandarin Chinese in few shots, multilingual adaptation is nevertheless effective, and
MSE and CER of a 2k-sample single-language model are achieved using 10\% or 15\% data on th-th and zh-cn respectively.
To conclude, though insufficient for few-shot learning on these languages, our approach is still powerful.

With all these results, we show that the general capability of the TTS task can be acquired by a single multilingual Byte2Speech model, and might be easily applied to a new language with few data, making a few-shot spoken language learner. More complete results are given in Appendix Section~B.

\subsection{Comparative studies}
\label{ablation}

\textbf{Diversity or Quantity}
By adding extra tiers of sources, the training data expand. We question if this matters.
Therefore, we experiment with downsampled data in each tier: we create a \textbf{T2-} dataset with T1 and T2 languages but only with the size of T1, and a \textbf{T3-} dataset with T2 and T3 languages but of T2 size. We use T2- or T3- as the source dataset, and then adapt the model to Greek. As shown in Figure~\ref{plot1}, T2- adaptation has a performance close to T2 while much better than T1, and it is similar for T3-. Therefore, for adaptation the diversity of the source corpus outweighs the data quantity.

\textbf{Diversity or Similarity} 
By intuition, adaptation to a target can be best aided by a similar source. This is also supported by the fact that T2 (with zh-hk) greatly helps zh-cn adaptation. Therefore, we examine it by co-training targets only using a closely related source, that is en-us for en-in, and Italian (it-it) for ro-ro. As indicated by the corresponding curves in Figure~\ref{plot1}, with only similar sources, results are close to T1. Cases are similar when co-training zh-cn with zh-hk. Therefore, the key to adaptation is not only leveraging a similar language but a combination of a diverse set of languages.

\textbf{Progressive or Direct}
Tier-wise progressive training plays an important role: we carry on \textbf{T3D} experiments that directly use full corpus. As a result, the model has worse source performance and fails on zh-hk. More, T3D source models show inferior Greek adaptation, as shown in Figure~\ref{plot3}. 

\textbf{When and How to Adapt} 
Instead of fine-tuning a well-trained base model, we start adaptation far before convergence on source languages. Starting from a semi-mature model (such as T2 500k steps or T3 700k steps) is beneficial: As shown in \textbf{T3 (650k)} results of Figure~\ref{plot3}, directly adding Greek to T3 results in a performance drop. While a mature network like \textbf{T3 (800k)} may lose its flexibility and fail on 10 samples. This is more phenomenal on T2, with up to 29\% absolute CER gap between different times to adapt. Therefore, a semi-mature network is beneficial for adaptation in our setting. Besides, we notice that at a few-shot regime the model tends to overfit. Therefore, we tune the proportion of the target by setting $p_i=0.1$, enforcing extra regularization from other languages. As demonstrated by the \textbf{T3 (0.1)} points for ro-ro and el-gr in Figure~\ref{plot1} and Figure~\ref{plot3}, few-shot performances are improved.

Additional details and omitted figures on ablation studies are available at Appendix Section~B.

\subsection{Extensibility to language expertise}

Although we aim at using minimal language-specific resources, the framework can be extended. For example, considering that under few-shot regimes some pronunciagtion rules never appear in the samples, we investigate Greek orthography and define a minimal set of 41 graphemes to represent Greek G2P rules. We then create pangram training sets sized 10 or 15 covering the set. 
As in results labeled as \textbf{Pan} in Figure~\ref{plot3}, significant benefits on few-shot cases are shown, particularly on 10-sample cases with 19\% relative CER reduction. Combined with $p_i=0.1$, the \textbf{Pan (0.1)} model reaches 37\% reduction.
Another approach is to augment inputs. We transform zh-cn texts into the Romanized Pinyin.
As shown in Figure~\ref{plot3}, zh-cn (Pinyin) obtains performance similar to other phonetic scripts, with CER close to toplines and the ground truth in 100 samples. If we further transform the script to mimic the Romanization of vi-vn (Vietnamese) in T3 to aid knowledge transfer, the \textbf{VietP} model gets extra improvements on few-shot regimes.
To conclude, the framework is flexible to improve on a particular language if we know more about it. See Appendix Section~A for details.

\begin{table*}[t]
\caption{Mean opinion score with 95\% confidence interval on different voices. For adaptation to target languages, we report low and extremely low (XLow) resource results, with the number of samples of each condition indicated in the following $\#$ line. We also give results on full-data single-language models, and the native name of languages to give an impression on input scripts.}
\label{mos}
\begin{center}
\begin{small}
\begin{sc}
\begin{tabular}{llllll}
\toprule
Language & en-us & de-de & zh-hk & vi-vn & te-in\\
Native name & English (US) & Deutsch & \includegraphics[height=1em]{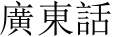} & \includegraphics[height=1em]{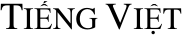} & \includegraphics[height=1em]{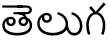}\\
\midrule
Recording & 4.27$\pm$0.13 & 3.95$\pm$0.13 & 4.01$\pm$0.13 & 4.59$\pm$0.08 & 4.52$\pm$0.11\\
Phoneme & 3.54$\pm$0.10 & 3.67$\pm$0.09 & 3.83$\pm$0.10 & 4.18$\pm$0.08 & 4.31$\pm$0.08\\
Byte & 3.69$\pm$0.10 & 3.44$\pm$0.10 & 3.76$\pm$0.10 & 4.15$\pm$0.08 & 4.30$\pm$0.09\\
\midrule
 & en-in & ro-ro & el-gr & th-th & zh-cn (pinyin)\\
 & Indian English & Română &\includegraphics[height=0.75em]{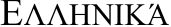} & \includegraphics[height=1em]{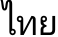} & han4-yü3 \\
\midrule
Recording & 4.72$\pm$0.08 & 4.71$\pm$0.09 & 4.48$\pm$0.11 & 4.65$\pm$0.09 & 3.82$\pm$0.12\\
Phoneme & 4.48$\pm$0.08 & 4.16$\pm$0.10 & 4.20$\pm$0.10 & 4.08$\pm$0.09 & 3.34$\pm$0.09\\
Byte Single & 4.56$\pm$0.08 & 4.38$\pm$0.08 & 4.41$\pm$0.08  & 4.18$\pm$0.10 & 3.58$\pm$0.09\\
\hspace{0.2cm} $\#$ & 9k & 7k & 7k & 7k & 19k \\
Byte & \\
\hspace{0.2cm} Low & 4.59$\pm$0.06 & 4.37$\pm$0.09 & 4.29$\pm$0.08 & 4.24$\pm$0.08 & 3.46$\pm$0.09\\
\hspace{0.4cm} $\#$ & 1k & 1k & 500 & 2k & 1k \\
\hspace{0.2cm} XLow & 4.37$\pm$0.09 & 3.74$\pm$0.10 & 3.30$\pm$0.12 & 3.75$\pm$0.11 & 2.75$\pm$0.09\\
\hspace{0.4cm} $\#$ & 30 & 30 & 30 & 500 & 100 \\
\bottomrule
\end{tabular}
\end{sc}
\end{small}
\end{center}
\end{table*}

\begin{table}[t]
\caption{Word level intelligibility (\%) on Romanian and Greek.}
\label{intel}
\begin{center}
\begin{small}
\begin{sc}
\begin{tabular}{lcccr}
\toprule
Language & Phoneme & Byte (10 samples) & Byte (30 samples)\\
\midrule
ro-ro & 99.9 & 99.4 & 99.3\\
el-gr & 98.9 & 93.5 & 96.8\\
\bottomrule
\end{tabular}
\end{sc}
\end{small}
\end{center}
\vskip -0.3in
\end{table}

\subsection{Subjective evaluations}
We perform subjective tests to evaluate our methods more accurately. As for naturalness, mean opinion score (MOS) results are given in Table~\ref{mos}. For source languages, byte models show comparable or only slightly worse results from the phoneme-based model. Hence multilingual byte models may produce natural speech on rich-resource source languages. On target languages, we choose to test performances on a low-resource case (\textsc{Byte Low}), along with an extremely low or few-shot case (\textsc{Byte XLow}) that shows good intelligibility, in comparison with single-language full-data byte models. We report Pinyin result as low-resource zh-cn are full of errors and thus less meaningful. Byte XLow has some gap compared to Byte Low models, while the Byte Low models show comparable MOS on en-in, ro-ro, and th-th, and only slight regression on el-gr and zh-cn, compared to \textsc{Byte Single}, using only 5\% to 29\% data. Furthermore, both Byte Low and Byte Single models have MOS comparable to phoneme models on all target languages. Results are consistent with objective metrics, showing our exceptional performance on low-resource adaptation.

We also test subjective intelligibility on Romanian and Greek, using the 30-sample  T3 adaptation model and the best 10-sample model, that is Pan (0.1) for Greek and T3 (0.1) for Romanian. As in Table~\ref{intel}, few-shot Greek models show good intelligibility with $>$90\% words correct, and for Romanian few-shot models are sufficient to produce close to 100\% intelligibility in tests. The results are highly consistent with CER, indicating that CER is a reliable metric for perceptual intelligibility.

\section{Model mechanisms}
From Google's multilingual translation to multilingual BERT, ML and NLP researchers have discovered that various languages can be well handled in a single shared model, which is counter-intuitive and different from typical multi-task learning using separate task-specific modules. Hence, the topic is valuable for general ML researchers: how a single model reaches such versatility on diverse tasks?

Inspired by the findings that there are both language-specific and -agnostic components in multilingual BERT embeddings \citep{DBLP:journals/corr/abs-1911-03310}, we believe that the answer may lie in the relationship between the multilingual model and the language-specific models. 
Therefore, we attempt to identify parameters or sub-models in the multilingual network that are important for each single language, and to explore the relationship between these particular sub-models.

We adopt data-driven pruning with Taylor criterions \citep{DBLP:conf/iclr/MolchanovTKAK17}, intending to identify salient neurons given input data. A neuron that is salient on data of a language might be important for the model to handle the language. The criterion is to consider the impact on the loss $\mathcal{L}$ when zeroing each neuron $h_i$ under an independent assumption, given training examples $\mathcal{D}$ of the language:
\[ \Delta \mathcal{L}(h_i) = |\mathcal{L}(h_i=0, D) - \mathcal{L}(h_i, D)| \]
, which is estimated using the Taylor expansion of $\Delta \mathcal{L}(h_i)$ near $h_i=0$. With a first-order expansion,
\[ \mathcal{L}(h_i=0, D) = \mathcal{L}(h_i, D) - \frac{\partial \mathcal{L}}{\partial h_i} h_i + R_1(h_i=0) \]

Ignoring $R_1$, the impact or saliency is approximated as 
\[ \Theta (h_i) = | \Delta \mathcal{L}(h_i)| = |\frac{\partial \mathcal{L}}{\partial h_i} h_i| \]

, which could be calculated in standard backpropagation.

We randomly sample 1k training examples on each language for saliency computation, on top of the T3 adaptation model to zh-cn with 2k samples. As we use a transformer, we compute saliency after each feedforward layer (plus non-linearity if present). Therefore, the impact of zeroing a neuron is equivalent to zeroing the corresponding parameters. Since layers are shared between steps in the input/output sequence, the saliency of each neuron is obtained by max-pooling across steps and averaging between samples. We then sort neurons for each layer and prune the lower half, assuming the remained network of 50\% parameters to be a language-specific model for each language. Next, for each pair of languages, we compare their specific sub-models by inspecting the proportion of neurons overlapped in both of them. Besides, we randomly split data of each language into halves and perform pruning separately, and then determine their overlappings in the same way.

\begin{figure}[t]
\centerline{\includegraphics[width=0.9\columnwidth]{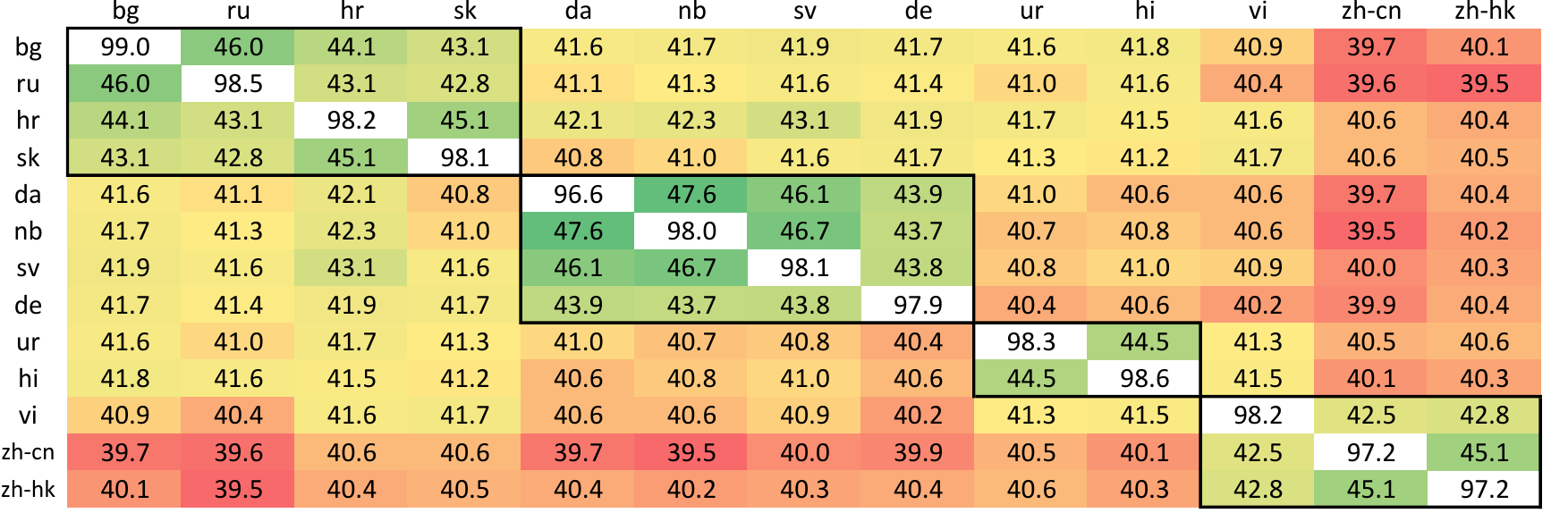}}
\caption{Percentage of overlapping neurons between selected language-specific models from post-ReLU activations in encoder layer 5. Diagonal values show intra-language overlappings.}
\label{sal}
\end{figure}

As shown in Figure~\ref{sal}, for each pair around 40\% neurons are salient on both languages, and the ratios correlate with language similarities. While intra-language overlappings are above 90\%, hence differences of salient neurons are due to differences of languages but not samples. Besides, as results are from a deep encoder layer, it not a direct result of distinct language IDs or character sets. Although we present those from encoder layer 5, such patterns appear in most layers.

Intuitively, languages with shared scripts have significant overlappings, such as bg/Bulgarian and ru/Russian from different branches of Slavic languages but both using Cyrillic alphabets. While the spoken form similarity matters and those with greater lexical, phonetic, or phylogenetic similarity are more overlapped, as shown by the closely related Germanic languages, especially the Nordic branch (da/Danish, nb/Norwegian, sv/Swedish), while de/German is farther. This even applies to languages with different scripts, such as ur/Urdu and hi/Hindi, the same spoken language in two different scripts. It is similar on distinct languages, such as vi (which is not phylogenetically related to but has rich loanwords from Chinese), zh-hk
, and the adaptation target zh-cn. Hence, during adaptation, the structures and parameters from similar source languages might be employed by the target. The phylogenetic relations can also be observed in the overlappings from bg to other Slavic languages, like hr/Croatian and sk/Slovak both using the Latin alphabet: hr has more overlapping as bg and hr are both South Slavic languages while sk is a more distant West Slavic language. 

We further verify our findings by actually pruning a T3 model with saliency computed from German, and then retraining it to monolingual models of T2 languages under low-resource settings, and next comparing the results of pruning with the language itself. As shown by Table~\ref{retrain}, the more a language is different from German, the greater performance drop by German-pruning can be observed. While if the model is randomly pruned, it will mostly fail. This indicates that our criterion correctly identifies neurons specific to German, and a similar language may better utilize the German sub-model.

All these evidences support that our model captures various high-level relations between languages just like multilingual BERT \citep{DBLP:conf/coling/RamaBE20}. Therefore, unlike previous works that require the manual design of language-specific and -agnostic modules such as per-language encoders \citep{DBLP:conf/icassp/NachmaniW19,DBLP:conf/interspeech/NekvindaD20}, we show that an architecture of language-specific sub-networks with partially shared parameters \textit{emerges} from the multilingual training of a single model, and the model may discover and utilize language similarities for both source languages and the transfer to targets to handle the task of multilingual TTS and low-resource adaptation.

\begin{table}[t]
\caption{MSE, CER, and the corresponding relative drop when pruning with German or the target language itself (\textsc{Self}), and then retraining on the target language.}
\label{retrain}
\begin{center}
\begin{small}
\begin{sc}
\begin{tabular}{llcccccc}

\toprule
& Target & Dutch & Polish & Russian & Korean & Arabics & Cantonese \\
\midrule

\multirow{3}{3em}{MSE}
&Self   & 0.458 & 0.510  & 0.466   & 0.451  & 0.520   & 0.541     \\
&German & 0.459 & 0.512  & 0.474   & 0.470  & 0.544   & 0.837     \\
&Drop (\%)       & 0.3\% & 0.4\%  & 1.8\%   & 4.2\%  & 4.6\%   & 54.9\%    \\
       \midrule
\multirow{3}{3em}{CER}
&Self   & 3.0\% & 2.1\%  & 5.2\%   & 13.2\% & 7.4\%   & 26.5\%    \\
&German & 3.0\% & 2.2\%  & 6.3\%   & 17.3\% & 10.8\%  & 67.4\%    \\
&Drop (\%)        & 0.5\% & 5.8\%  & 21.4\%  & 30.9\% & 47.3\%  & 154.2\%  \\
      \bottomrule
\end{tabular}
\end{sc}
\end{small}
\end{center}
\end{table}

\section{Related work}
Various prior works attempt to build multilingual neural TTS models.
\citet{DBLP:conf/icassp/LiZSWC19} is a closely related work that proposed byte-based speech recognition and synthesis model. However, the work was done only on few languages and did not touch the low-resource scenarios. Other multilingual methods introduced adversarial training \citep{DBLP:conf/interspeech/ZhangWZWCSJRR19}, or
per-language encoder \citep{DBLP:conf/icassp/NachmaniW19,DBLP:conf/interspeech/KorteKK20}, possibly with parameters predicted from language IDs \citep{DBLP:conf/interspeech/NekvindaD20}. 
Multilingual TTS helps low-resource languages to a great extent, both in multilingual training and adaptation \citep{DBLP:conf/interspeech/LiZ16,DBLP:conf/sltu/DemirsahinJG18,DBLP:conf/interspeech/BaljekarRB18,DBLP:conf/interspeech/KorteKK20}. Our framework is developed upon the phoneme-based language-balanced multilingual transformer TTS \citep{DBLP:conf/interspeech/YangH20} which extend the method to 40+ languages, while we remove the needs of phoneme-inputs, outperform their performance on low-resource adaptation, and explore the few-shot regime. Various more complex methods were proposed to better leverage transfer learning, pretraining, and semi-supervised learning, such as using pretrained text embeddings and self-supervised decoder pretraining \citep{DBLP:conf/icassp/ChungWHZS19}, and fine-tuning from an autoencoder with discrete latents trained on unpaired target speeches \citep{DBLP:conf/interspeech/ZhangL20, DBLP:conf/icassp/LiuTLL20}. \cite{DBLP:conf/icml/RenTQZZL19} applies dual learning between recognition and synthesis, along with denoising autoencoding, and \cite{DBLP:conf/kdd/XuTRQLZL20LR} further added rich-resource pretraining and knowledge distillation. Data augmentation with injected noise \citep{DBLP:conf/interspeech/LiuWLC20} also helps. Besides, transfer from other languages could be obtained by language expertise such as designing a unified set of phonemes \citep{DBLP:journals/corr/abs-2005-10441}, often derived from IPA \citep{DBLP:conf/interspeech/ChenTYL19,DBLP:conf/sltu/DemirsahinJG18}, or creating feature vectors by phonology \citep{DBLP:conf/interspeech/StaibTTMFLG20}. While our method eliminates the need of these language-specific expertise and complicated pipelines using auxilliary models and data.

\section{Conclusions and future work}
We present a systematic approach to build a multilingual Byte2Speech TTS model and show that it is capable to match phoneme-based performance on both standard and low-resource adaptation scenarios. We also deepen our understanding of the mechanism by a novel interpretation. Future work will focus on improving few-shot performances and further exploring the model mechanism.


\bibliographystyle{plainnat}
\bibliography{neurips_2021}


\appendix

\section{Implementation details}
\subsection{Model}
We follow the best setting of \citet{DBLP:conf/aaai/Li0LZL19} to use a transformer with scaled sinusoidal position encoding, 6 encoder layers, 6 decoder layers, and 8 heads, along with a 5-layer convolutional post-net, except that encoder pre-net is not adopted. We follow the parameters used in Tacotron2 \citep{DBLP:conf/icassp/ShenPWSJYCZWRSA18Tacotron2} to create acoustic features of mel spectrograms. The processed mel spectrograms are used to train a WaveNet under the setting of \citet{DBLP:conf/aaai/Li0LZL19}, to generate waveforms used in subjective tests. 

As for training, we apply Adam optimizer with exponential learning rate decay from 1e-3 to 1e-5 in 850k steps using a decay rate of 0.01, which is reset at tier transitions and the beginning of adaptations. We use dynamic batching with at most 8000 output frames in each batch \citep{DBLP:conf/aaai/Li0LZL19}, and train with data parallelism on four V100 GPUs. The timings for tier transitions and adaptation are found by grid search according to the results on source languages and the target language respectively. On the source languages, the model uses total 1M steps to reach convergence, taking 6.3 days. The best result among the training steps are reported for each metric. On target languages the time varies, depending on the difficulty of the language and the number of samples. Starting from T3, a 10-sample ro-ro model takes 100k steps to reach best CER, while a 1k-sample ro-ro model takes 300k steps, and a 1k-sample zh-cn one using characters takes around 400k steps. 

We follow the approach in \citet{DBLP:conf/interspeech/YangH20} to add language embeddings and speaker embeddings from the corresponding MLPs after the encoder. Therefore, with byte inputs the language identity is not directly given to the encoder. We also attempted to inject the language embeddings at the input side, either by concatenating with input embeddings, or by special tokens \citep{DBLP:journals/tacl/JohnsonSLKWCTVW17}. We also attempted to use deeper (e.g. 16-layer) transformers.
However, in experiments we found that it leads to similar or even worse performance on source languages.

To facilitate reproduction, an implementation based on open codes and datasets will be available at \url{https://github.com/mutiann/few-shot-transformer-tts}, although due to differences of dataset the results are not identical to what we report in the paper on proprietary datasets.

\subsection{Comparative studies}
In comparative studies, we determine the training steps for alternative training settings by comparing the training loss to our standard setting. Starting from 30k steps we train the model with T1 languages in T2- until it reaches the loss of 350k steps of the standard model, and then add the rest languages in T2- until it reaches the loss of 500k steps. Similarly, we perform experiments on T3- dataset starting from the standard T1 at 350k steps. For T3D adaptation, we start from the step when reaching training loss identical to the standard T3 at 700k steps.

\subsection{Extensibility to language expertise}
As for Greek, we find a minimal set of 41 graphemes of letters and multigraphs to cover major G2P rules. In the randomly selected 10 and 15 samples, seven rarest elements of the minimal set never appear, while all of them appear in the evaluation set, making errors inevitable. Therefore, by using such language-specific expertise, we train on an alternative set of 10 and 15 samples, each forming a pangram when viewed together, covering all the elements in the minimal set. Other conditions are set identical to the standard T3. We pick the pangram sets with total script lengths close to the original 10- and 15-sample-set (with difference $<$10\%), to avoid the impact of different sample lengths. 

As for Mandarin Chinese, we transform scripts into Pinyin by lexicon, with words segmented and tones labeled by numbers. We obtain performance similar to other phonetic scripts. However, the gap is large on few-shot scenarios, potentially due to phonetic differences to most source languages. We observe that errors mostly occur on tones and pronunciation of the vowel /ü/ absent in the standard Latin alphabet. Therefore, we further turn to encourage transfer from Vietnamese in T3, which has similar tones to Chinese but uses a Latin alphabet with particular diacritics to denote tones. We adopt the diacritics and substitute /ü/ with /\includegraphics[height=0.6em]{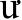}/ in Vietnamese to create the VietP model, and reach better performance. The full results of the experiments are given in Figure~7.

\subsection{Model mechanisms}

We choose to inspect the language-specific sub-models with 50\% neurons pruned. On some layers (such as the ReLU outputs on the last encoder layer) due to the sparsity of ReLU activation only less than half of the neurons are used, making all the overlapping proportions 100\%. This can be avoided if we choose a larger pruning ratio such as 75\%. However under such a large ratio, the pruned model is too small to capture many complicated TTS tasks in the retraining experiment, leading to failures of retraining on zh-hk and ko-kr. Hence we choose a smaller ratio of 50\%.

In retraining experiments, since a 50\% pruned model still possess the flexibility to learn a language if sufficient data are given, we use a low-resource scenario to better analyze the result. All except zh-hk use 500 samples, while zh-hk uses 2k samples, which is the minimum size we found to produce an intelligible self-pruning model. Experiments are performed on a converged T3 model of 1M steps.

\section{Additional experimental results}

\begin{figure*}[t]
\vskip 0.2in
\centerline{\includegraphics[width=\textwidth]{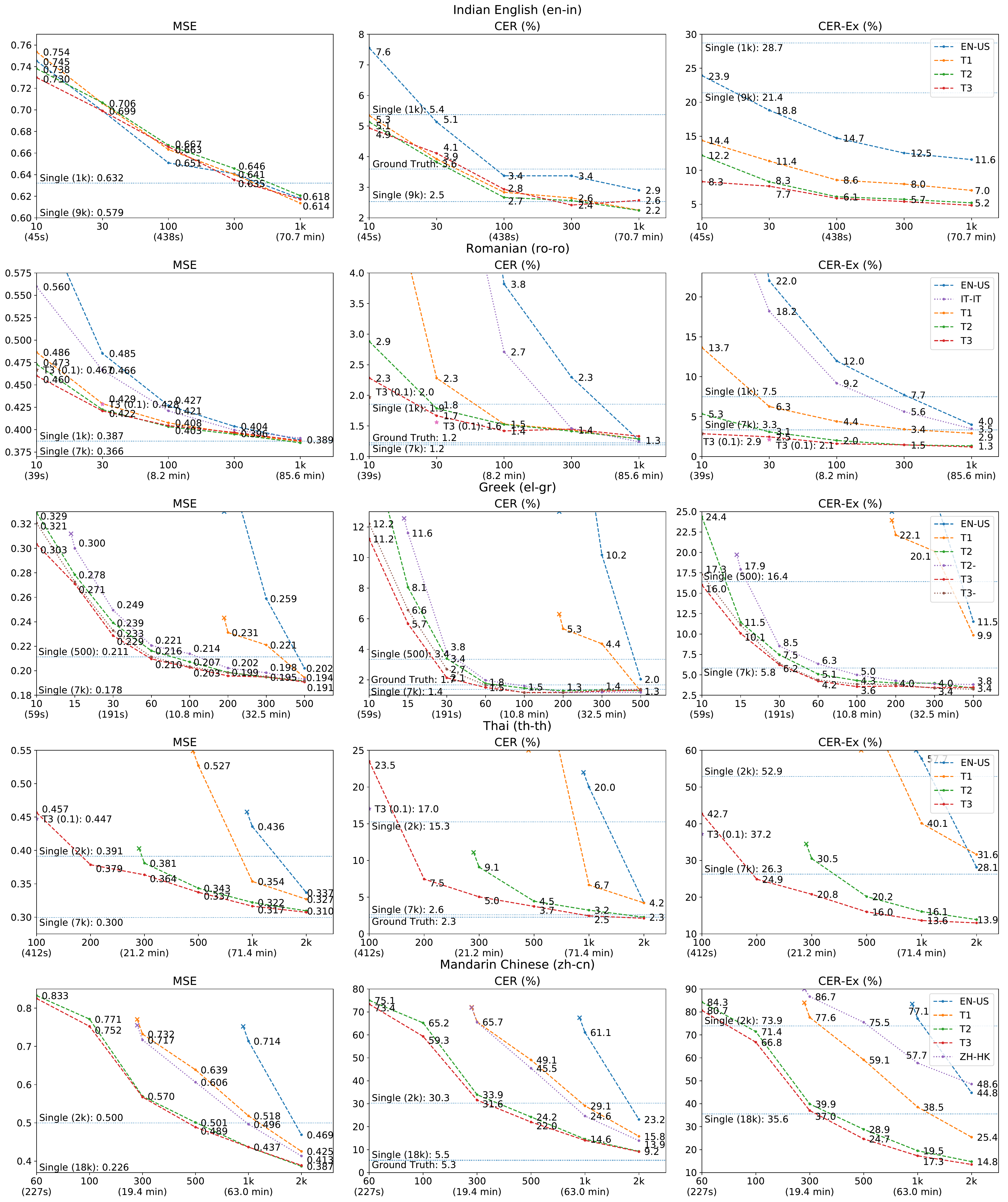}}
\caption{Objective results for Indian English (en-in), Romanian (ro-ro), Greek (el-gr), Thai (th-th), and Mandarin Chinese (zh-cn).}
\label{plot_all}
\end{figure*}

\subsection{Adaptation}
The full results of MSE, CER, and CER-Ex for en-in, ro-ro, and el-gr are given in Figure~6. which are consistent with our findings given in the paper:
\begin{itemize}
  \item MSE, CER, and CER-Ex similar to single-language models can be reached with much fewer data.
  \item Good intelligibility can be achieved in few shots.
  \item The more source languages, the better performance, especially when there are limited data.
  \item Better robustness than single-language models as shown by CER-Ex.
  \item Results on different metrics are consistent to each other.
\end{itemize}

The results for th-th and zh-cn are also given in Figure~6. zh-cn is more difficult and even with 2k samples the best model could still not reach the ground truth CER, while T2 with zh-hk helps a lot. Nevertheless, the results are consistent with our findings, except for few-shot language acquisition.

\subsection{Comparative studies}

The full results for \textbf{diversity or quantity} on Greek are given in Figure~6, showing consistency with our findings, so are the results for \textbf{diversity of similarity} (added by the zh-hk curve on zh-cn), and for \textbf{when and how to adapt} (as supported by Figure~7).

 As for the question of \textbf{progressive or direct}, the T3D source model has similar or worse source CER: 2.46\% on en-us, 1.26\% on de-de, 47.82\% on zh-hk, and 9.65\% on te-in; and it has much worse Greek adaptation results, as in Figure~7. Besides, the full results of Pangram and Pinyin experiments are consistent with findings in the paper as well. Details for implementation are mentioned in Section A.

\begin{figure*}[t]
\vskip 0.2in
\centerline{\includegraphics[width=\textwidth]{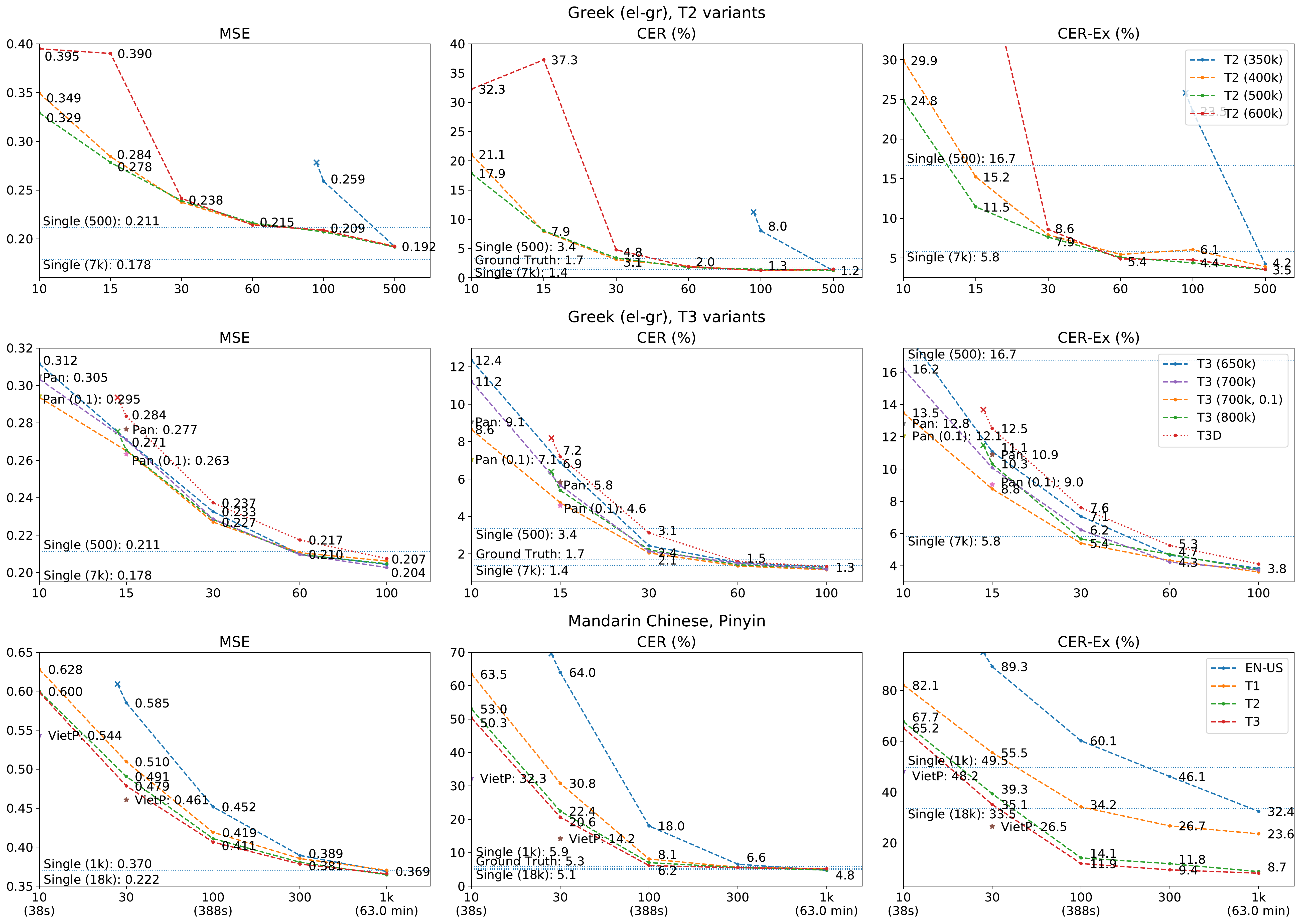}}
\caption{Objective results for alternative settings in Greek (el-gr) and Mandarin Chinese (zh-cn) using Pinyin.}
\label{plot_all_ex}
\end{figure*}

\subsection{Cross-lingual speaker transfer}
Although not our focus, our model also supports cross-lingual speaker transfer similar to previous works \cite{DBLP:conf/interspeech/YangH20}, that is to transfer a speaker's voice to another language. Audio samples are available on our webpage, showing the cross-language transfer on a Japanese female speaker and an American male speaker. As shown by the results, the model could generate English speech in the Japanese speaker's voice, and Japanese speech in the American speaker's voice, even though such cases are unseen during training. We further compare with the recordings for corresponding English utterances by the Japanese speaker, and find that by our model more natural English speech with less Japanese accent could be produced.

\subsection{Multi-speaker target language data}
All adaptation targets in our experiments are datasets with thousands of samples from a single professional speaker. However, in real world scenarios it is often more economical to collect a dataset on the target language with multiple non-professional speakers, each contributing hundreds of samples, as in the settings and datasets of \cite{DBLP:conf/lrec/HeCKRKGDJJSP20}. Therefore we also perform low-resource adaptation on their open Gujarati (gu-in) datasets released under CC-BY-SA 4.0, with 4k samples from 35 speakers. Under the settings identical to our previous T3 adaptation experiments, we could reach a CER of 6.84\%, showing that our method is applicable to the scenario as well.

\section{Dataset details}

Details of our dataset is given in Table~\ref{dataset}. Besides, We demonstrate the linguistic traits of languages involved in Figure~8, including their writing systems and phylogenetical relationships. As shown in the figure, our full dataset cover a wide variety of spoken languages and writing systems, but the lower tiers (such as T1 marked in green) are much narrower, and our target languages (marked in red) are selected that each of them shares different level of similarity with the source languages when adapted from each tier.

\begin{figure*}[t]
\vskip 0.2in
\label{langtraits}
\centerline{\includegraphics[width=\textwidth]{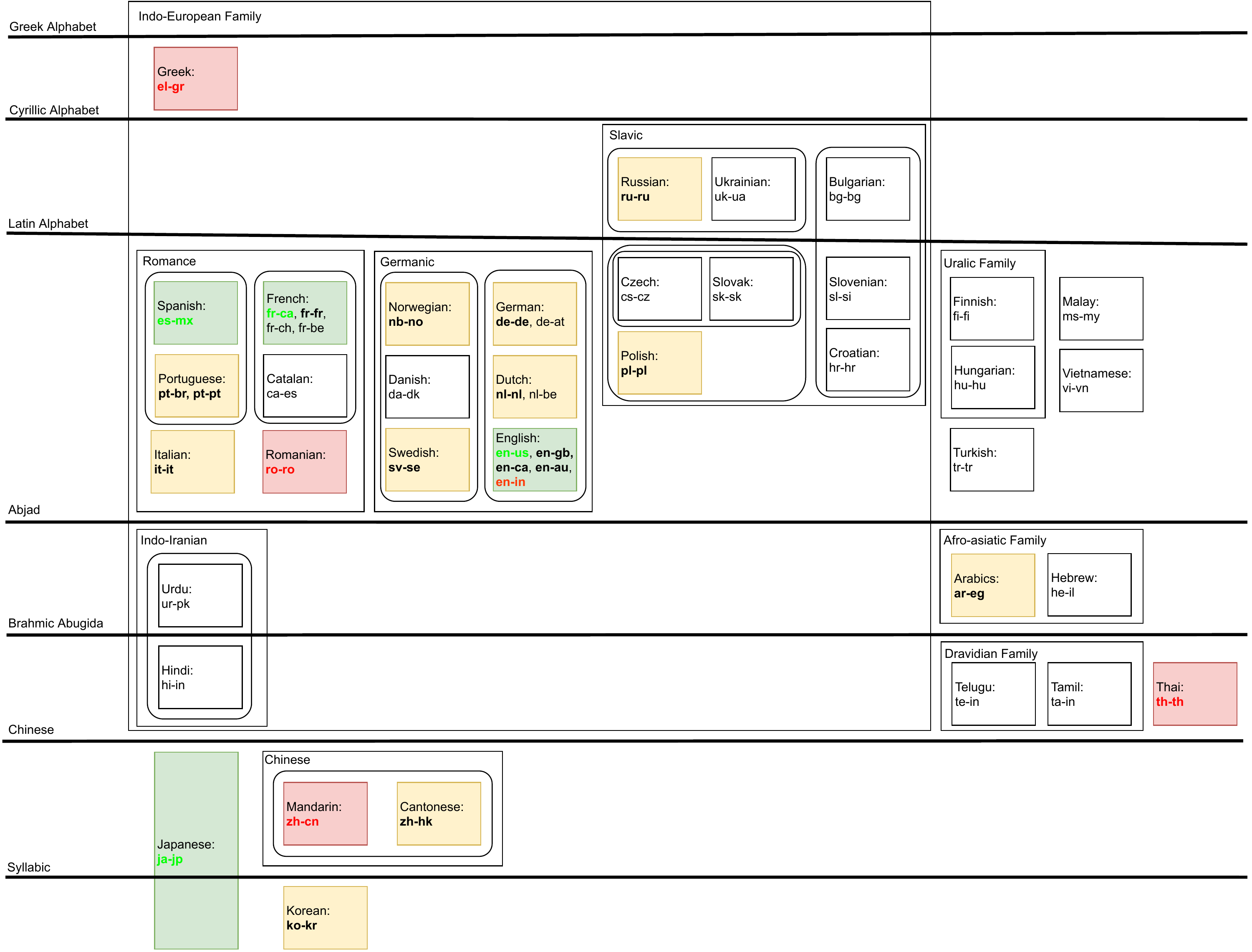}}
\caption{Linguistic traits of the involved languages. T1 languages are given in green bold codes in green boxes, T2 in bold codes in yellow boxes, target languages in red codes in red boxes, and the rest are T3 languages. Writing systems of the languages are indicated. Language families and their direct branches are shown in rectangular boxes to group the languages. Languages with additional similarity are grouped in more fine-grained rounded boxes.}
\end{figure*}

\begin{table}[t]
\caption{List of languages and their codes of each tier in our experiment.}
\label{dataset}
\begin{center}
\begin{small}

\begin{tabular}{llll}
\toprule
Code & Language & Code & Language \\
\midrule
\multicolumn{2}{c}{\textsc{Tier 1}}            & da-dk & Danish (Denmark)                  \\
en-us & English (US)              & de-at & German (Austria)                  \\
es-mx & Spanish (Mexico)          & de-ch & German (Swiss)                    \\
fr-ca & French (Canada)           & fi-fi & Finnish (Finland)                 \\
ja-jp & Japanese (Japan)          & fr-be & French (Belgium)                  \\
\multicolumn{2}{c}{\textsc{Tier 2}}            & fr-ch & French (Swiss)                    \\
ar-eg & Arabics (Egypt)           & he-il & Hebrew (Israel)                   \\
de-de & German (Germany)          & hi-in & Hindi (India)                     \\
en-au & English (Australia)       & hr-hr & Croatian (Croatia)                \\
en-ca & English (Canada)          & hu-hu & Hungarian (Hungary)               \\
en-gb & English (UK)              & ms-my & Malay (Malaysia)                  \\
fr-fr & French (France)           & nl-be & Dutch (Belgium)                   \\
it-it & Italian (Italy)           & sk-sk & Slovak (Slovak)                   \\
ko-kr & Korean (Korea)            & sl-si & Slovenian (Slovenia)              \\
nb-no & Norwegian Bokmal (Norway) & ta-in & Tamil (India)                     \\
nl-nl & Dutch (Netherlands)       & te-in & Telugu (India)                    \\
pl-pl & Polish (Poland)           & tr-tr & Turkish (Turkey)                  \\
pt-br & Portuguese (Brazil)       & uk-ua & Ukrainian (Ukraine)               \\
pt-pt & Portuguese (Portugal)     & ur-pk & Urdu (Pakistan)                   \\
ru-ru & Russian (Russia)          & vi-vn & Vietnamese (Vietnam)              \\
sv-se & Swedish (Sweden)          & \multicolumn{2}{c}{\textsc{Targets}}               \\
zh-hk & Cantonese (Hong Kong)     & el-gr & Greek (Greece)                    \\
\multicolumn{2}{c}{\textsc{Tier 3}}            & en-in & English (India)                   \\
bg-bg & Bulgarian (Bulgaria)      & ro-ro & Romanian (Romania)                \\
ca-es & Catalan (Spain)           & th-th & Thai (Thailand)                   \\
cs-cz & Czech (Czech)             & zh-cn & Mandarin Chinese (Mainland China) \\
\bottomrule

\end{tabular}

\end{small}
\end{center}
\end{table}


\end{document}